\def\year{2020}\relax
\documentclass[letterpaper]{article} 
\usepackage{aaai20}  
\usepackage{times}  
\usepackage{helvet} 
\usepackage{courier}  
\usepackage[hyphens]{url}  
\usepackage{graphicx} 
\usepackage{graphicx}  
\usepackage{times}
\usepackage{amsmath,epsfig}
\usepackage{graphicx,color,amssymb,graphics,amsmath,subfigure,amsfonts,epsfig,setspace,graphics}
\usepackage[lined,boxed,commentsnumbered, ruled]{algorithm2e}
\usepackage{ragged2e}
\usepackage{url}
\usepackage{times}
\usepackage{epsfig}
\usepackage{bm}
\usepackage{epstopdf}
\usepackage{subfigure}

\title{Towards Omni-Supervised Face Alignment for Large Scale Unlabeled Videos}
\author{Congcong Zhu\textsuperscript{\S\dag}, Hao Liu\textsuperscript{\dag\ddag}\thanks{Corresponding Author is Hao Liu.}, Zhenhua Yu\textsuperscript{\dag\ddag} and Xuehong Sun\textsuperscript{\dag\ddag}\\ 
\textsuperscript{\dag} School of Information Engineering, Ningxia University, Yinchuan, 750021, China\\ 
\textsuperscript{\ddag} Collaborative Innovation Center for Ningxia Big Data and Artificial Intelligence \\Co-founded by Ningxia Municipality and Ministry of Education, Yinchuan, 750021, China\\
\textsuperscript{\S} School of Computer Engineering and Science, Shanghai University, Shanghai, 200444, China
\\ congcong.zhu\_nxu@outlook.com, \{liuhao, zhyu, sunxh\}@nxu.edu.cn
}
 
\begin{document}

\maketitle

\begin{abstract}
	In this paper, we propose a spatial-temporal relational reasoning networks (STRRN) approach to investigate the problem of omni-supervised face alignment in videos. Unlike existing fully supervised methods which rely on numerous annotations by hand, our learner exploits large scale unlabeled videos plus available labeled data to generate auxiliary plausible training annotations. Motivated by the fact that neighbouring facial landmarks are usually  correlated  and coherent  across consecutive frames, our approach automatically reasons about discriminative spatial-temporal relationships among landmarks for stable face tracking. Specifically, we carefully develop an interpretable and efficient network module, which disentangles facial geometry relationship for every static frame and simultaneously enforces the bi-directional cycle-consistency across adjacent frames, thus allowing the modeling of intrinsic spatial-temporal relations from raw face sequences. Extensive experimental results demonstrate that our approach surpasses the performance of most fully supervised state-of-the-arts.
\end{abstract}

	\section{Introduction}
\label{sec:intro}

Face alignment~(\textit{a.k.a.}, facial landmark detection) aims to detect multiple facial landmarks for the given facial image or video sequence, which has dominated a crucial step in many facial analysis tasks such as  face identification and recognition~\cite{HuLT14,DBLP:journals/tcsv/LiuLFZ19} and face animation~\cite{DBLP:conf/cvpr/RothTL15,DBLP:conf/eccv/LiuZZL16}. While significant works have been devoted to face alignment recently~\cite{DBLP:journals/tpami/Liu19,wayne2018lab,kumar2018disentangling}, the practical performance hardly meets the precise requirements in the scenario of performing large scale unlabeled videos. The major reasons are two-fold: 1) Existing methods intensely rely on the sheer volume of training annotations. 2) It is tedious to annotate the spatial-temporal structures on the massive set of frames. Hence, we are supposed to propose an accurate and robust face alignment algorithm to automatically annotate the large amounts of unlabeled face videos. 

\begin{figure} [tb]
	\centering
	\begin{minipage}[h]{\linewidth}
		\subfigure[Spatial Dendritic Relational Constraint]{
			\centering
			\includegraphics[width=0.95\textwidth]{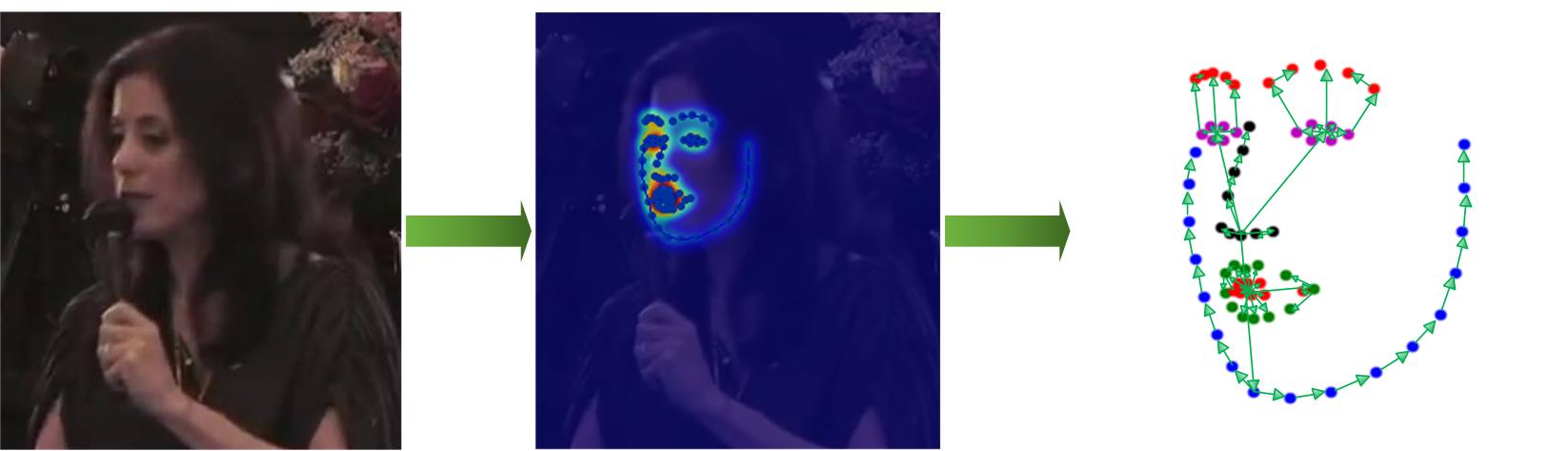}.jpg}
	\end{minipage}%
	
	\begin{minipage}[h]{\linewidth}
		\subfigure[Temporal Cycle-Consistent Relation]{
			\centering
			\includegraphics[width=0.95\textwidth]{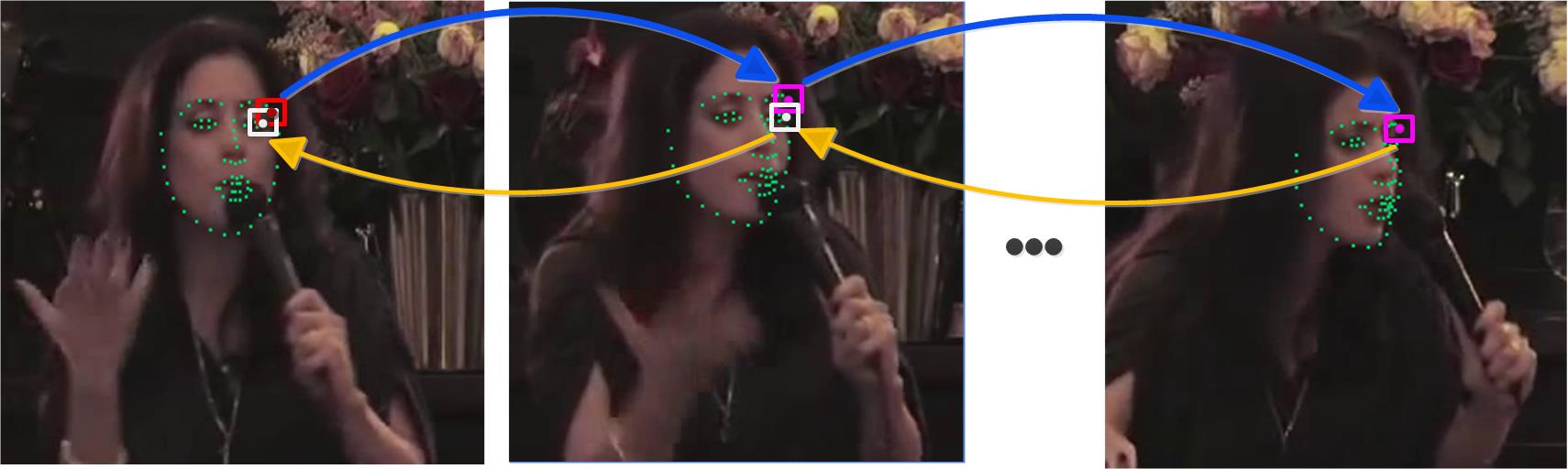}}
	\end{minipage}
	\caption{Insight of our proposed STRRN. Our method processes the large scale unlabeled video by automatically distilling the spatial-temporal knowledge, \textit{i.e.}, a) exploiting the facial geometry relationship by holding a dendritic structure,  b) enforcing the bi-directional cycle-consistency temporally on consecutive frames. We leveraged these self-supervised spatial-temporal relation to generate extra reliable training annotations for model update.}\label{fig:insight}
\end{figure}

Throughout recent literatures, face alignment is dominated by the regression-based approaches~\cite{Trigeorgis_2016_CVPR,wayne2018lab,kumar2018disentangling,DBLP:conf/eccv/GuoLZ18,DBLP:journals/tpami/Liu19,DBLP:journals/pami/LiuLFZ18}, which typically seek discriminative feature-to-shape mappings with preserving the shape constraint. Methods such as  LAB~\cite{wayne2018lab} and PCD-CNN~\cite{kumar2018disentangling} impose the relative geometry among multiple landmarks, making the plug-in module  pose-invariant for accurate performance. However, these fully-supervised methods require large amounts of precise hand-crafted annotations for training and cannot be directly adopted for unlabeled data. Besides the spatial relationship, video-based methods such as CCR~\cite{DBLP:conf/eccv/Sanchez-LozanoM16} and TSTN~\cite{DBLP:journals/pami/LiuLFZ18} enforces the consistency relation temporally across adjacent frames, which improves the stableness versus the jitter problem during tracking. 
One major issue in this type of method is that the training efficiency still relies on the tremendous volume of per-frame annotations, where it is challenging to manually annotate numerous frames. 
To address this issue, we investigate into the omni-supervised learning towards face alignment inspired  by~\cite{DBLP:conf/cvpr/RadosavovicDGGH18,DBLP:conf/cvpr/DongYWW0S18}. Our basic idea aims at reasoning disentangling spatial-temporal relationship by exploiting large scale unlabeled videos plus all available training annotations. The self-reasoned spatial-temporal relation allows us to trust their predictions on the unseen video data. The intuitive idea is presented in Fig.~\ref{fig:insight}.

\begin{figure*} [tb]
	\centering
	\includegraphics[width=0.82\linewidth]{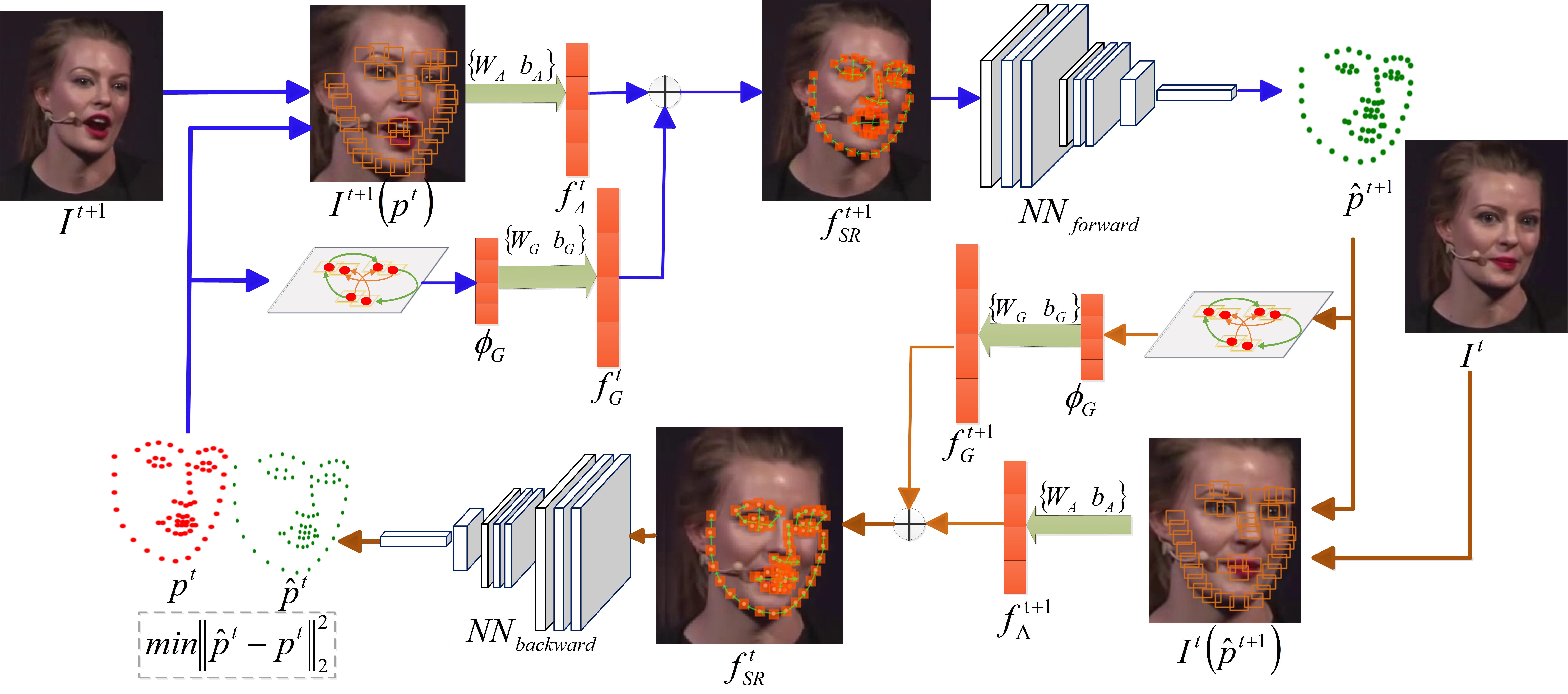}
	\caption{Network design of our STRRN. Overall, our method processes in a bi-directional tracking routing, which is visualized in the blue arrow~(forward) and brown arrow~(backward).  For each routing, our architecture incorporates with the spatial and temporal modules. Accordingly, the spatial module is fed with the current frame and initialized landmarks~(tracked results of previous frame). Then the module reasons about both the spatial appearance and geometry relationship, resulting discriminative features with preserving a dendritic-structure. Having obtained these features, we directly feed them to a neural networks~(NN) structure to track landmarks in the temporal module.   The network parameters of both directions are shared, thus allowing the temporal modeling of cycle-consistent relation for stabling tracking. During learning phase, the STRRN is self-supervised based on the outcomes of the backbone detector. This figure is best viewed in color pdf file. }\label{fig:STRRN}
\end{figure*}

To tackle the omni-supervised face alignment, we propose a spatial-temporal relational reasoning networks~(STRRN) approach, which automatically distills interpretable knowledge of large scale unlabeled video data. 
To produce plausible training annotations for model update, our approach reasons about meaningful relations of unlabeled videos in both the spatial and temporal dimensions accordingly.  For modeling of the spatial relation, we first divide the whole face into different semantic components~(\textit{e.g.}, eyes, nose, mouth, facial cheek and etc.), where the nose is assumed as the root. Then our STRRN  disentangles the component-based appearance and geometric information with preserving a dendritic structure. For modeling of the temporal relation, we ensure that our model tracks forward until the target and it should arrive the starting position in the backward order.
This principally enforces the cycle-consistent temporal relation on consecutive frames for reliable tracking.  
To learn the network parameters, we adopt a cooperative and competitive strategy to exploit the complementary information from both the tracking module and the backbone detector. Fig.~\ref{fig:STRRN} shows the detailed network architecture of the proposed STRRN. 
To evaluate the effectiveness of our proposed approach, we carry out extensive experiments on folds of large scale unlabeled video datasets and experimental results indicate compelling performance versus state-of-the-art methods.

\section{Related Work}
\label{sec:related}
We briefly review some related literatures of existing face alignment methods and knowledge distillation models.

\textbf{Conventional Face Alignment:} Existing face alignment methods are roughly classified into image-based~\cite{cao2012face,XiongSDM,ZhuLLT15,SunCVPR14} and video-based~\cite{DBLP:conf/eccv/Sanchez-LozanoM16,Tzimiropoulos15}. Earlier shallow models intend to seek a sequence of discriminative linear feature-to-shape mappings, so that the initialized shape is adjusted to the target one in a coarse-to-fine manner. Representative cascade regression-based methods include explicit shape regression~(ESR)~\cite{cao2012face}, supervised descent method~(SDM)~\cite{XiongSDM} and coarse-to-fine shape searching~(CFSS)~\cite{ZhuLLT15}.
To make the image-based methods suitable for video data, one common solution is to regard the outcomes of previous frames as initializations for the following frames via a tracking-by-detection method~\cite{DBLP:journals/pami/WangYZL15}. However, this method could only extract 2D spatial appearances from static images and cannot explicitly exploit the temporal information on consecutive frames. 
To circumvent this problem, video-based methods~\cite{DBLP:conf/eccv/Sanchez-LozanoM16,Tzimiropoulos15,Khan_2017_ICCV} learn to memorize and flow the temporal consistency information across frames, which improves the robustness to the jitter problem in visual tracking. 
However, one major issue in these methods is that these linear mappings are too weak to exploit the complex and nonlinear relationship between image pixels and shape variations in unconstrained environments.

\textbf{Deep Neural Networks for Face Alignment:} These years have witnessed that deep learning~\cite{KrizhevskySH12,0483bd9444a348c8b59d54a190839ec9} contributes breakthroughs in face alignment~\cite{DBLP:conf/eccv/GuoLZ18,Trigeorgis_2016_CVPR,wayne2018lab,kumar2018disentangling,SunDCNC13,ZhangCFAN14,Jourabloo_2017_ICCV,DBLP:journals/tpami/Liu19,DBLP:journals/pami/LiuLFZ18,DBLP:conf/eccv/PengFWM16}, which achieves  improvements by taking  advantages of nonlinear regression and end-to-end feature learning.
Representative deep learning-based methods include  mnemonic descent method~(MDM)~\cite{Trigeorgis_2016_CVPR}, two-stream transformer networks~\cite{DBLP:journals/pami/LiuLFZ18}, pose conditioned dendritic convolutional neural networks~(PCD-CNN)~\cite{kumar2018disentangling}, HourGlass Network~(HGN)~\cite{DBLP:conf/eccv/NewellYD16} and a look-at-boundary~(LAB) method~\cite{wayne2018lab}. 
However, most methods are performed by fully supervised learning and likely give rise to the upper-bounded performance with existing labeled data~\cite{DBLP:conf/cvpr/RadosavovicDGGH18}.  Moreover, it is difficult to manually annotate the massive set of adjacent frames in practice. More recently, reinforcement learning~\cite{DBLP:journals/ml/Williams92} is introduced to face alignment recently~\cite{DBLP:journals/tpami/Liu19,DBLP:conf/eccv/GuoLZ18}, which teaches a policy to select actions sequentially by
obtaining a greater feedback value each time. The same sense has been discussed in~\cite{XiongT15} that cascade regression can be considered as a simple and nature case of imitation learning.  In contrast to prior efforts, our approach aims to reason about meaningful knowledge from amounts of unlabeled data. Meanwhile, both the spatial geometry relationship and temporal dependency information are simultaneously exploited in a unified deep learning architecture. As a result, our architecture serves as a scalable solution for processing large scale unlabeled data source. 

\textbf{Knowledge Distillation Models:}  Knowledge distillation targets on making predictions on unlabeled data and further using them to update the model, which has been adopted in various scenarios such as visual detection~\cite{DBLP:conf/cvpr/RadosavovicDGGH18,DBLP:conf/apsipa/ChouCC18,DBLP:conf/cvpr/ChuOLW16} and model compression~\cite{DBLP:journals/corr/RomeroBKCGB14,DBLP:conf/cvpr/GuptaHM16}.  For example, Romero \textit{et al.}~\cite{DBLP:conf/cvpr/GuptaHM16} adopts a shallow and wider teacher model to learn a thin and deep student model. Without training a large set of models, Radosavovic \textit{et al.}~\cite{DBLP:conf/cvpr/RadosavovicDGGH18} presented a data distillation method to inferring predictions from unlabeled data by multiple transformations, resulting new training annotations. 
Our work is mostly related to a supervised-by-registration~(SBR)~\cite{DBLP:conf/cvpr/DongYWW0S18} method where the detector's predictions on unlabeled data  are leveraged to train itself. 
However, SBR~\cite{DBLP:conf/cvpr/DongYWW0S18} ignores the spatial geometric relationship of neighbouring landmarks, thus likely making bias predictions during tracking. On contrast, in our approach, we automatically reason the meaningful spatial-temporal knowledge from large scale unlabeled videos by following the success of the relation networks framework~\cite{DBLP:conf/cvpr/HuGZDW18}. As a result, the dendritic geometry in the spatial dimension and cycle-consistency in the temporal dimension are simultaneously exploited to generate reliable new annotations.

\section{Learning Spatial-Temporal Relational Reasoning Networks}
\label{sec:method}

In our approach, the main goal aims to address the omni-supervised face alignment for large scale unlabeled videos. The basic idea is to distill spatial-temporal knowledge from numerous amounts of unlabeled frames, thus generating extra reasonable training annotations. To achieve this, we propose a spatial-temporal relational reasoning networks~(STRRN) architecture which consists of the spatial and temporal modules. Specifically, our architecture typically reasons about tree-structural relationship spatially based on the appearance and geometry features. Temporally, we enforce the bi-directional cycle-consistency relation across consecutive frames for robust landmark tracking.

Fig.~\ref{fig:STRRN} demonstrates the network design which incorporates with the bi-directional executions of the STRRN.
Suppose we have an downloaded unlabeled video set $\mathcal{D}=\{\bm{I}_i^t\}^{1:T}_{1:N}$   with $N$ video clips, where  $\{\bm{I}_i\}^{1:T}= \{\bm{I}^{1}_i,\bm{I}^{2}_i,...,\bm{I}^{t}_i,...,\bm{I}^{T}_i\}$ denotes each face sequence that consists of $T$ frames. It is valuable to stress that we assume that each face frame to be tracked is already cropped by the learned face detector~\cite{DBLP:journals/spl/ZhangZLQ16}. 
Let $\bm{p}^t={[(x_1,y_1),(x_2,y_2),\cdots,(x_L,y_L)]^t} \in \mathbb{R}^{2L \times 1}$ denote the coordinates of the facial shape vector for the $t$th frame, where $L$ specifies the landmark quality within the whole face~(e.g., 68 landmarks in 300-W~\cite{DBLP:journals/ivc/SagonasATZP16}). 
For notation brevity, we assume there is only one unlabeled video with $T$ frames and thus ignore the video index $i$.
The main objective of face tracking in videos is to transform face sequence $\bm{I}^t$~(a set of local patches cropped based on the previous outcomes)  to the shape coordinates $\bm{p}^{t}$ at time $t$.
One alternative to model the shape coordinates is to regressing a set of heat maps by feeding the entire face image like~\cite{kumar2018disentangling,wayne2018lab}, which indicates the superiority on promoting the alignment performance. However, face tracking in videos is always sensitive to the detection. That is, the low-quality detections probably cause the drifting issue for tracking results. How to achieve the synergy between detection and tracking is investigated in~\cite{Khan_2017_ICCV,DBLP:conf/eccv/GuoLZ18}, which appears beyond the range of this work.   Next, we describe the network specification for the spatial and temporal modules accordingly in our proposed STRRN. Moreover, we present the detailed training procedure.

\textbf{Spatial Relational Reasoning Module:}
It has been studied that conditioning  structural information of neighbouring landmarks provides much potential to promote the performance~\cite{BelhumeurJKK11,DBLP:journals/tip/LiuLFZ17,kumar2018disentangling}.
Inspired by this, the motivation of our spatial module aims at disentangling a dendritic geometry relationship based on the tracked results of precious frame and then tracking on the current face frame.
To exploit the geometry relation,  we first divide the whole landmarks $\bm{p}$ into $C$ components using the tracked results of previous frame, \textit{e.g.}, facial cheek, left eye, right eye, nose, mouth, by strictly following the standard of the 300-W annotations\footnote{\url{https://ibug.doc.ic.ac.uk/resources/300-W/}}. Based on the initialized landmarks, we extract a set of  raw patches and divide them to $C$ groups. Let  $\bm{I}^t(\bm{p}^t_c,d)$ represent raw patches  cropped from the $t$-th frame via the initialized shape vector $\bm p$, where $c$ denotes the group identity up to $C$ and the patch size is assigned to $d$~(ignored for simplicity). 
Within each group, we compute the coupled \textit{appearance feature} and \textit{geometry feature} via deep neural networks.
Specifically, we embed the raw patches to the immediate features in each group as~(ignoring the patch index):
\begin{eqnarray}  \label{equ:appearance}
\bm{f}^t_A= \sigma(\bm{W}_A \cdot \bm{I}^{t}(\bm {p}^{t-1})+\bm{b}_A),
\end{eqnarray}
where the parameters $\{\bm{W}_A,\bm{b}_A\}$ specify the spatial appearance weights and $\sigma(\cdot)$ is the nonlinear function, \textit{e.g.}, ReLU~\cite{KrizhevskySH12}, respectively. 

For the description of the geometry relation, we compute the relation feature by the comparisons of the paired $(m,n)$  patches within each group. To make it invariant to affine transformations due to different poses, we assume the nose as the base location for robust feature learning. More specifically, we compute the geometry features $\left(\bm{\phi}_G^m,\bm{\phi}_G^n\right)$~(simply ignoring the time-stamp index)  of the paired  patches  by concatenating the relative distance vector $[ log(|x^m-x^n|),log(|y^m-y^n|)]$ and the pose-invariant vector $[{log(|x^m-x^*|),log(|y^m-y^*|)}]$~($x^*$ and $y^*$ specify the horizontal and vertical coordinates of the nose, respectively). 
To exploit the complexly nonlinear spatial relationship, we embed these geometric features into the high-dimensional  feature representations as 
\begin{eqnarray}  \label{equ:hidden}
\bm{f}^t_G(m)= \sigma(\bm{W}_G  \cdot \text{CONCAT}[\bm{\phi}_G^m, \bm{\phi}_G^n]+\bm{b}_G),
\end{eqnarray}
where $\{\bm{W}_G,\bm{b}_G\}$ are the spatial geometric weights and $\text{CONCAT}(\cdot)$ denotes the concatenation operator.
For each group, we perform  the spatial relation features via addition,
\begin{eqnarray}  \label{equ:geometry}
\bm{f}^t_{\text{SR}}= \text{CONCAT}[\bm{f}^t_G,\bm{f}^t_A(1),\cdots,\bm{f}^t_A(m)],
\end{eqnarray}
where $\bm{f}^t_G$ specifies the holistic geometric feature concatenation across all groups. Note that $\bm{f}^t_A(m)$ specifies the appearance feature for $m$-th group where $m$ is the landmark scalability for each group. 

Lastly, we concatenate all the embedded features into one long feature vector $\bm{f}_{\text{SR}}^t$ across $m$ groups for the $t$th frame. Furthermore, we feed the spatial relation feature $\bm{f}_{\text{SR}}$ into a tiny convolutional network to estimate landmark positions of following frames.

In summary, the spatial module learns the correlation of neighbouring landmarks by grouping them separately. By exploiting local appearance for each facial group, it guides landmarks moving towards more plausible prediction. Moreover, our module reasons with the nose-centered geometric relations, which enforces view-consistency for robust landmark tracking versus various view changes~\cite{DBLP:conf/eccv/ZavanNSBS16}.

\textbf{Temporal Relational Reasoning Module:} In the term of the temporal module, our goal is to reason about the bi-directional cycle-consistent relation between adjacent frames. In other words, the module ensures that our tracker proceeds forward along the video and it should arrive at the starting position in the backward order. Thus, this allows our model to reason with the cycle-consistency constraint of unlabeled videos for stabling temporal modeling~\cite{DBLP:conf/aaai/MeisterH018}. 
To achieve this, we learn two trackers parameterized by the designed deep neural networks, playing the video forward and backward accordingly. In order to minimize the discrepancy of tracked results in both forward and backward orders, we employ a cycle-consistency check function to evaluate the tracking reliability. 

As demonstrated in Fig.~\ref{fig:STRRN}, there are two steps for each tracker. First, we regard the tracked results $\bm{p}^t$ to perform the spatial relation feature~$\bm{f}_{\text{SR}}^{t+1}$ on $\bm{I}^{t+1}$ via the proposed spatial module. Sequentially, we feed the feature ~$\bm{f}_{\text{SR}}^{t+1}$ to a forward-NN to estimate the landmarks $\hat{\bm{p}}^{t+1}$ on the ${(t+1)}$th frame.
In the similar manner on the $t$th frame $\bm{I}^t$, we employ the tracked landmarks $\bm{p}^{t+1}$ as the initialization to perform the spatial relation feature~$\bm{f}_{\text{SR}}^{t}$. Having obtained $\bm{f}_{\text{SR}}^{t}$, we pass it to the backward-NN and achieve the feedback results $\hat{\bm{p}}^{t}$. Both the forward and backward tracking executions are formulated as follows:
\begin{eqnarray} \label{equ:tracing}
\hat{\bm{p}}^{t+1} = \text{NN}_{\text{backward}}(\bm{f}^{t+1}_{\text{SR}}), \quad \hat{\bm{p}}^{t} = \text{NN}_{\text{forward}}(\bm{f}^{t}_{\text{SR}}),
\end{eqnarray}
where $ \text{NN}_{\text{forward}}(\cdot)$ and $ \text{NN}_{\text{backward}}(\cdot)$ denote the neural networks with multiple layers, aiming to estimate the coordinates of the landmarks. Moreover,  the parameters both forward and backward networks are shared. 
To  make the tracking process efficient, we leveraged a squeezed deep architecture which is equipped with only a few layers of convolutions, max pooling, nonlinear ReLU activation and fully connection. 

To further evaluate the reliability of the cyclic execution, we compute the cycle-consistency check function, which ensures that the tracked landmarks should be re-traced back again to the start. By doing this, our premise is that if the tracked results of the forward-tracker are plausible, the landmarks should return to the same positions after tracking in the backward order.  In this way, our method achieves reliable tracking results with the bi-directional cycle-consistency check temporally across adjacent frames. The check loss is computed as follows:
\begin{eqnarray} \label{equ:cc_loss}
\mathcal{L} = \| \hat{\bm{p}}^{t} - {\bm{p}}^{t} \|_2^2,
\end{eqnarray}
where $\|\cdot\|$ denotes the $\ell^2$ norm to measure the discrepancy between the tracked and feedback results.

\textbf{Training Procedure:}
In our approach, we investigate into the omni-supervised setting of face alignment particularly for large scale unlabeled video data. Targeting on omni-supervised learning, our method learns to trust the predictions across the massive amounts of frames and moreover includes these proxy predictions as training annotations. 
Since our tracker module may not always succeed, the predictions are not directly considered as the reliable training supervision signals. 
Hence, to make the prediction robust for the model update, we leverage a backbone detector, \textit{e.g.}, MDM~\cite{Trigeorgis_2016_CVPR} and HGN~\cite{DBLP:conf/eccv/NewellYD16}, learned by amounts of annotated images, to evaluate the tracking reliability. With the detector, our tracker is iteratively refined by the reliable reasoned training data itself, offering much potential to surpass the performance of existing fully-supervised approaches. 

\begin{figure} [tb]
	\centering
	\includegraphics[width=\linewidth]{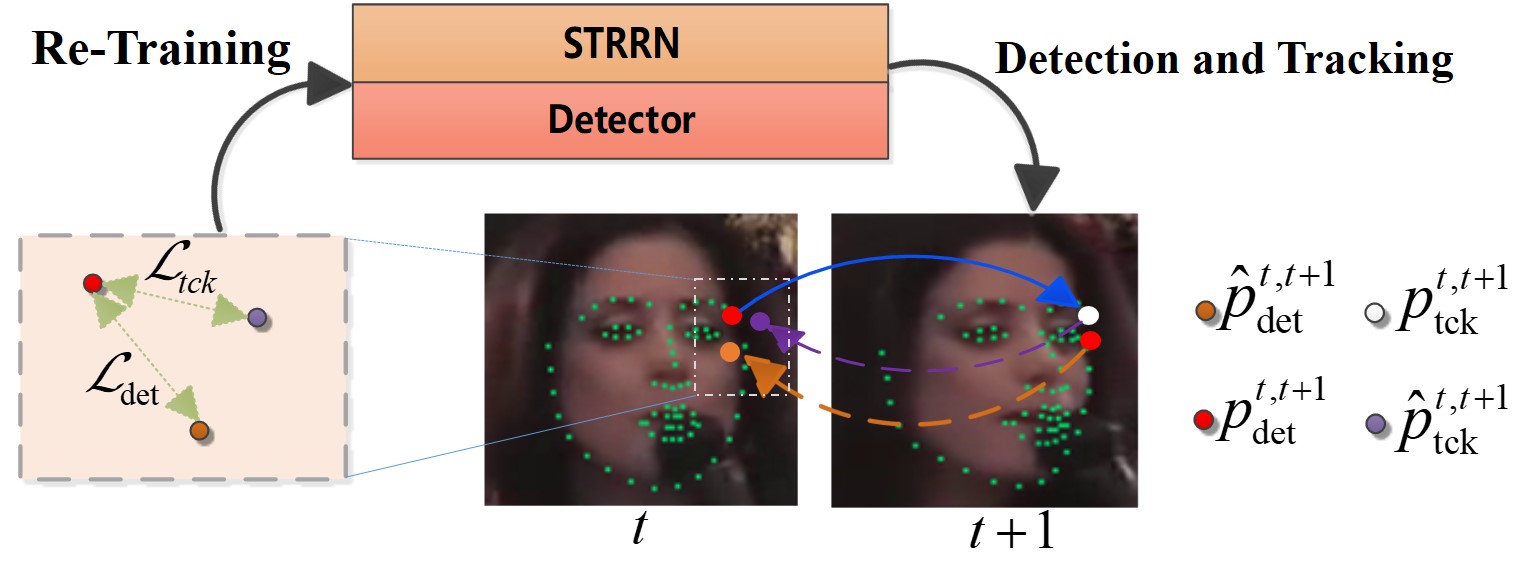}
	\caption{Training procedure of our STRRN. Basically, there are two complementary steps: 1) We pass frames onto our STRRN by distilling the spatial-temporal relation and perform the tracked results. 2) We utilize the backbone detector to evaluate the tracked results.  Consequently, our tracker is self-supervised by justifying the reliability of either the cycle-consistency and the detected results. This really allows our model to trust these reasoned predictions for robust and efficient training. }\label{fig:training}
\end{figure}

As illustrated in Fig.~\ref{fig:training}, our training procedure  justifies two-fold  scenarios: 1) The higher reliability will be given when the tracking results $\bm{p}^t_{\text{tck}}$ go back again to the start $\hat{\bm{p}}^{t-1}_{\text{tck}}$, where the discrepancy $\mathcal{L}_{\text{tck}}$ enforces the temporal continuity in our STRRN. 2) A negative feedback is provided when the discrepancy $\mathcal{L}_{\text{det}}$ between backward-tracked results $\hat{\bm{p}}^{t-1}_{\text{tck}}$ and those ${\bm{p}}^{t-1}_{\text{det}}$ generated by the backbone image-based detector. This means the performance degrades to the pre-trained backbone detector when we undergo imprecise and unreliable tracking results. 

To exploit the model update, we select the reliable outputs of both tracking and detection by minimize the complete loss functions of $\mathcal{L}_{\text{det}}$ and $\mathcal{L}_{\text{tck}}$ as follows:
\begin{eqnarray} \label{equ:loss}
\mathcal{L}_{\text{ensemble}} = \sum_{t=1}^{T}\mathcal{L}^t_{\text{tck}}+\lambda \sum_{t=1}^{T}\mathcal{L}^t_{\text{det}},
\end{eqnarray}
where $\lambda$ is the weight of the ensemble for both detected and tracked results.
With (\ref{equ:loss}), our method achieves to be self-supervised by reasoning with the reliable tracking. This principally benefits from the competitive and cooperative executions of our tracker and the backbone detector. Moreover, the complementary feedback of both parts leads to reliable new training annotations $\mathcal{D}_{\text{det}}$ and $\mathcal{D}_{\text{tck}}$, where $\mathcal{D}_{\text{det}}$  and $\mathcal{D}_{\text{tck}}$ denote the detected and tracked for module update. \textit{Algorithm}~\ref{alg:STRRN} shows the detailed pseudocode for training procedure.

\begin{algorithm}[tb] \label{alg:STRRN}
	\DontPrintSemicolon
	\LinesNumbered
	\KwIn{Training set: $\mathcal{D}=\{\bm{I}^t\}^{t=1:T}$~(ignoring sample $i$ for simplity) and $\mathcal{S}$~(samples from 300-W).}
	\KwOut{Network parameters: $\{\bm{W}_A, \bm{b}_A\}$, $\{\bm{W}_G, \bm{b}_G\}$, $\text{NN}_\text{forward}$ and $\text{NN}_\text{backward}$.}
	//Training the backbone Detector~(Both MDM and HGN were employed in the experiments);\\
	\For{$t\in \text{video}<T$}{
		/*Detecting on the $t$th frame:*/\\ 
		$\bm{p}_{\text{det}}^{t} = \text{Detector}(\bm{I}^t);$\\
		/*Forward tracking on the $t$th frame:*/
		$\bm{p}^{t}_{\text{tck}} = 
		\text{Forward-STRRN}(\bm{I}^{t}, \bm{p}^{t-1}_{\text{det}});$\\
		/*Backward tracking based on detected landmarks:*/
		$\hat{\bm{p}}^{t-1}_{\text{det}} = 
		\text{Backward-STRRN}(\bm{I}^{t-1}, \bm{p}^{t}_{\text{det}});$\\
		/*Backward tracking based on tracked landmarks:*/
		$\hat{\bm{p}}^{t-1}_{\text{tck}} = 
		\text{Backward-STRRN}(\bm{I}^{t-1}, \bm{p}^{t}_{\text{tck}});$\\
		/*Criticizing the tracking reliability by the detection according to (\ref{equ:cc_loss}):*/ \\
		$\mathcal{L}_{\text{det}}=\|\hat{\bm{p}}^{t-1}_{\text{det}}-{\bm{p}}^{t-1}_{\text{det}}\|_2^2$, $\mathcal{L}_{\text{tck}}=\|\hat{\bm{p}}^{t-1}_{\text{tck}}-{\bm{p}}^{t-1}_{\text{det}}\|_2^2$;\\
		/*Generating new training annoations by thresholding $\mathcal{T}$:*/ \\
		\If{$\mathcal{L}_{\text{det}},\mathcal{L}_{\text{tck}}< \mathcal{T}$}{\If{$\mathcal{L}_{\text{tck}}>\mathcal{L}_{\text{det}}$}{$\mathcal{D}_{\text{tck}} \leftarrow \mathcal{D}_{\text{tck}} \cup (\bm{I}^t, \bm{p}_{\text{det}}^t)$;} {$\mathcal{D}_{\text{det}} \leftarrow \mathcal{D}_{\text{det}} \cup (\bm{I}^t, \bm{p}_{\text{tck}}^t)$} ;}
	}
	/*Data distilling and model re-training;*/\\
	\quad Train(Detector, $\mathcal{D}_{\text{det}}\cup\mathcal{S}$);   \quad Train(STRRN,$\mathcal{D}_{\text{tck}}$);\\
	\textbf{Return}: $\{\bm{W}_A, \bm{b}_A\}$, $\{\bm{W}_G, \bm{b}_G\}$.
	\caption{Training Procedure of Our STRRN}
\end{algorithm}

\section{Experiments}
To justify the effectiveness of the proposed STRRN, we represent folds of experimental results and analysis based on three downloaded large scale video datasets.
\label{sec:exp}

\textbf{Evaluation Datasets}: \textit{\textit{300VW}~\cite{DBLP:conf/iccvw/ShenZCKTP15}}: The 300 Videos in the Wild (300VW)  Dataset was collected specific for video-based face alignment. The dataset contains 114 videos captured in various conditions and each video has around 25-30 images every second. By following \cite{DBLP:conf/iccvw/ShenZCKTP15}, we utilized 50 sequences for training  and  the remaining 64 sequences were used for testing.
The whole testing set was divided into three categories: well-lit, mild unconstrained and challenging. In particular, the Category Three exploits many difficult cases of face sequences, which highlights the superiority of the proposed approach.
It is valuable to note that we leveraged the 300-W~\cite{DBLP:journals/ivc/SagonasATZP16} training set to initialize our backbone detector. 	In our approach, we mainly improved the tracker module by exploiting large scale unlabeled videos, which is quite different from SBR~\cite{DBLP:conf/cvpr/DongYWW0S18} principally to reinforce the landmark detector. Hence, we do not justified the superiority of our method on the image-based datasets such as 300-W~\cite{DBLP:journals/ivc/SagonasATZP16}.

\textit{YouTube-Face~\cite{DBLP:conf/cvpr/WolfHM11}} and \textit{YouTube-Celebrities\cite{DBLP:conf/cvpr/KimKPR08}}: We also leveraged two large scale unlabeled video datasets including YouTube-Face~\cite{DBLP:conf/cvpr/WolfHM11} and YouTube-Celebrities\cite{DBLP:conf/cvpr/KimKPR08}  for evaluation. Specifically, YouTube-Face consists of 3425 video clips with 1595 person identities, while YouTube-Celebrities consists of videos with 35 celebrities which undergo large variations due to different poses, illumination and partial occlusions. By following the setting in SBR~\cite{DBLP:conf/cvpr/DongYWW0S18} which is also designed suitable to unlabeled videos , we filtered these videos with low resolution and trained our STRRN unsupervisedly.

\textbf{Experimental Setup:}
Following the evaluation protocol employed in~\cite{DBLP:journals/pami/LiuLFZ18}, we computed the root mean squared error~({RMSE}) normalized by the inter-pupil distance~\cite{XiongSDM,DBLP:journals/ivc/SagonasATZP16} and cumulative error distribution (CED) curves for comparisons in our experiments. To be specific, we performed the RMSEs of all frames within each category in 300VW and then averaged them as final performance. Moreover, we leveraged the CED curves of {RMSE} errors to quantitatively evaluate the performance. Besides, as discussed in MDM~\cite{Trigeorgis_2016_CVPR}, there is no consistent way of the normalized factors for alignment, \textit{e.g.}, inter-ocular distance~\cite{XiongSDM,DBLP:journals/ivc/SagonasATZP16} or inter-pupil distance~\cite{ZhuLLT15,cao2012face}. To clarify the results, we also computed the area-under-the-curve~(AUC) by respectively thresholding the RMSEs at 0.05 and 0.08. 

For each video to be aligned, we rescaled each face frame in the size of $[315\times 315]$. Our spatial relational reasoning module accepts $L$~raw patches, where $L$ is assigned 68 according to the 300-W annotation and each patch is rescaled in size of $30\times 30$.
For hyper-parameters in our STRRN, we empirically set the discounted factor $\lambda$ to 0.4 and the thresholding $\mathcal{T}$ to the normalized RMSE 0.02 during generating extra training annotations. More details will be made publicly in our release model and source code.

\begin{table} [tb] \small
	\caption{Comparisons of averaged errors of our proposed STRRN with the state-of-the-arts~(68-lms, in chronological ranking). \textit{It can be seen from the results that our unsupervised STRRN outperforms  both the unsupervised SBR and other fully supervised approaches. Moreover, the performance of both the backbone MDM and HGN is promoted by integrating with our STRRN.} 
	}
	\label{tab:sta}
	\vspace{5pt}\centering
	\begin{tabular}{ | l  | c | c | c  |} 
		\hline
		\textbf{Method} & \textbf{Cate-1} &\textbf{ Cate-2 }& \textbf{ Cate-3}   \\
		\hline
		
		CFSS~(2015)     & 7.68 & 6.42 & 13.67 \\
		TCDCN~(2015)   & 7.66 & 6.77 & 14.98  \\
		CCR~(2016)  & 7.26 & 5.89 & 15.74 \\
		iCCR~(2016)  & {6.71 }& {4.00} & {12.75}  \\
		MDM~(2016)    & 9.88 & 8.14 & 9.81\\
		HGN~(2016)  & 6.12 & 5.33 & 7.04\\
		{TSTN}~(2018)   &{{5.36}} & {{{4.51}}} & {{12.84}}  \\
		SBR~(2018) & 5.77 & 4.90 & 12.66\\
		\hline
		\textbf{STRRN (Backbone-MDM)}   &\textbf{5.03} & {\textbf{4.74}} & \textbf{6.63}  \\ 
		\textbf{STRRN (Backbone-HGN)}  & \textbf{5.31} & \textbf{4.77} & \textbf{8.06} \\
		$\dagger$~\textbf{Semi-STRRN}$^{-50}$   &\textbf{4.21} & {\textbf{4.18}} & \textbf{5.16}  \\ 
		$\dagger$~\textbf{Semi-STRRN}$^{-25}$   	&\textbf{4.49} & {\textbf{4.05}} & \textbf{5.88}  \\
		$\dagger$~\textbf{Semi-STRRN}$^{-10}$   &\textbf{4.67} & {\textbf{4.00}} & \textbf{5.93}  \\
		\hline
	\end{tabular}
	
	{  
		\begin{itemize} \small
			\item 	[	$\dagger$ ] ${-50}$, ${-25}$ and ${-10}$ denote the percentages of 300VW annotations  were employed for  our  STRRN training, respectively.
		\end{itemize}
		
	}
	
	
\end{table}	


\begin{figure*} [tb]
	\centering
	\begin{minipage}[h]{0.25\linewidth}	
		
		\subfigure[Averaged error comparisons]{
			\centering \label{fig:SBRa}
			\includegraphics[width=\textwidth,height=30mm]{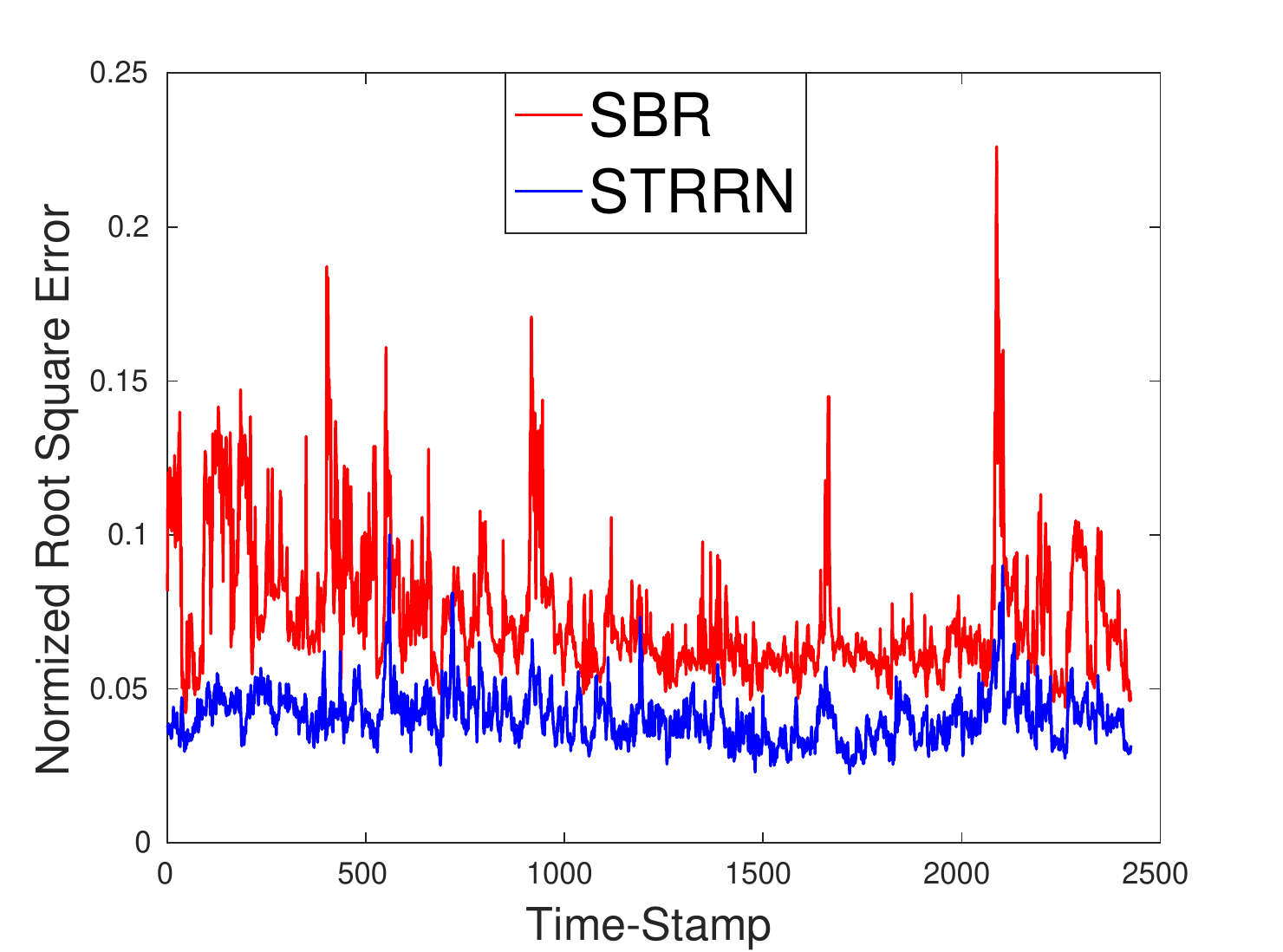}}
	\end{minipage}%
	\begin{minipage}[h]{0.65\linewidth}
		\subfigure[Representative resulting face sequence]{
			\centering \label{fig:SBRb}
			\includegraphics[width=\textwidth,height=30mm]{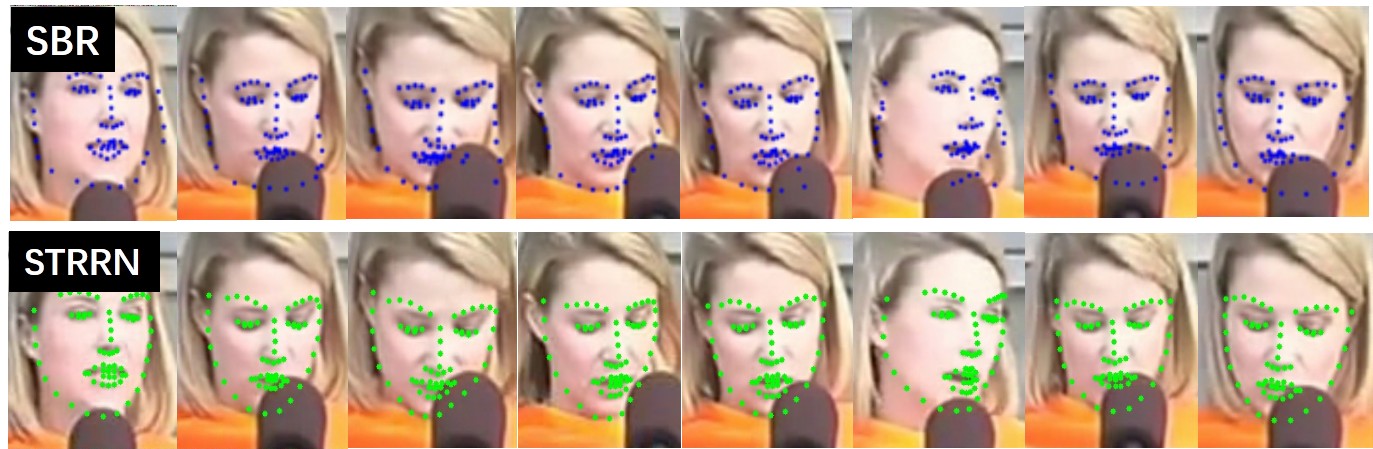}}
	\end{minipage}
	\caption{Comparisons of our STRRN and SBR on the challenging 517-th sequence of the 300VW Category Three. Both-fold results demonstrate that our model achieves low errors especially on occluded facial cheek thanks to the spatial constraint.}
	
\end{figure*}

\textbf{Comparisons with State-of-the-art Methods:}
We compared our proposed STRRN with image-based methods~\cite{XiongSDM,ZhuLLT15,ZhangPAMI15,Trigeorgis_2016_CVPR} and video-based methods~\cite{DBLP:conf/eccv/Sanchez-LozanoM16,DBLP:journals/pami/LiuLFZ18}, where these methods were fully supervised by large amounts of annotations.
For fair comparisons, we re-produced the results with their released codes or cropped the reported results from their original papers. Note that only our STRRN and SBR~\cite{DBLP:conf/cvpr/DongYWW0S18} only leveraged 300-W images for model training.

To investigate the effectiveness of our proposed approach, we compared our approach with the state-of-the-art face alignment methods, which were designed for both static images and tracking in videos via a fully-supervised manner.
For fair comparisons, we first discarded all annotations for model training by following the unsupervised setting.
Hence, we did not compare our STRRN with the newly-proposed supervised method dubbed FHR~\cite{tai-FHR-2019}.
To further highlight the advantage of our approach, we trained our model by only a subset of annotations, which is termed as \textit{Semi-STRRN} in  
Table~\ref{tab:sta}, which tabulates the comparisons of our method with the state-of-the-arts in both unsupervised and semi-supervised on the 300VW datasets. From these results, we see that our proposed STRRN significantly outperforms other face alignment methods by a large margin especially on the challenging 300VW Category Three.  This is mainly because our cycle-consistent execution exploits more interpretable cues to learn discriminative spatial-temporal features for robust face tracking. Moreover, seeing from Table~\ref{tab:sta} which tabulates the comparisons of STRRN versus semi-STRRN, we found that even with weakly-labeled supervision signals, we still perform very compelling results on the challenging cases including large poses, diverse expressions and severe occlusions. This also demonstrates the effectiveness of the proposed spatial-temporal relation modeling of the bidirectional orders, where these learned temporal relations are helpful to reinforce the tracking stabilization. 

Lastly, we investigated the effectiveness of the temporal relational reasoning module in our STRRN.  We used SBR~\cite{DBLP:conf/cvpr/DongYWW0S18} as the baseline method. Both our method and SBR~\cite{DBLP:conf/cvpr/DongYWW0S18} were learned on 300VW without any supervision, and tested on the selected 517-th face sequence of the most challenging category.
Fig.~\ref{fig:SBRa} shows the sequential errors of our STRRN and SBR~\cite{DBLP:conf/cvpr/DongYWW0S18} across a series of time stamps.
From these curves, we achieve a clear and stable results by a large margin compared with SBR~\cite{DBLP:conf/cvpr/DongYWW0S18}, which shows the stabilization of ours versus various temporal motions. This is mainly because the testing sequence undergo continuous occlusion over time steps. Clearly, our STRRN achieves to reason about the robust spatial-temporal feature representation across frames for stabling landmark tracking, while SBR~\cite{DBLP:conf/cvpr/DongYWW0S18} always degrades especially on the facial cheek.  The similar sense is visualized on the qualitative comparisons in Fig.~\ref{fig:SBRb}, where our STRRN achieves the coherency cues in contrast to SBR~\cite{DBLP:conf/cvpr/DongYWW0S18}.

\begin{table} [tb] \small
	
	\caption{Comparisons of STRRN on different training datasets~(AUC (\%) thresholding RMSEs at 0.05 and 0.08).}
	\vspace{5pt}\centering
	\begin{tabular}{|l | c c  |}
		\hline
		\textbf{Video Dataset}   & \textbf{AUC@0.05}&\textbf{AUC@0.08} \\
		\hline 
		{300VW} & 26.67 & 78.07 \\
		{300VW+YouTubeCelebrities } &28.31 & 77.24\\
		{300VW+YouTubeFace} & {36.33} & {80.43}\\
		\textbf{All Combined} & \textbf{39.48} & \textbf{85.06}\\
		\hline
	\end{tabular}\label{tab:sta_auc}
\end{table}


\textbf{Investigation on Unlabeled Video Datasets:}
We further compared our proposed STRRN on different datasets including {300VW}~\cite{DBLP:conf/iccvw/ShenZCKTP15},
YouTube-Face~\cite{DBLP:conf/cvpr/WolfHM11} and
YouTube-Celebrities\cite{DBLP:conf/cvpr/KimKPR08}. By carefully following the employed setting in SBR~\cite{DBLP:conf/cvpr/DongYWW0S18}, we regarded our STRRN trained on the 300VW dataset without any annotations as the baseline method, where 68 landmarks were employed for evaluation. 
Even without any every-frame annotations, our method trained with either of large scale unlabeled video data outperforms that trained with only 300VW.
Moreover, Table~\ref{tab:sta_auc} tabulates the AUC comparisons  at the thresholding RMSEs at 0.05 and 0.08  on different datasets. 
From Table~\ref{tab:sta_auc}, the outperformed results prove that we succeed in reasoning the meaningful spatial-temporal knowledge from unlabeled videos, thus making reliable extra  annotations for model update.


\textbf{Ablation Study:} 
To further investigate the ablation experiments of the proposed components in our STRRN, we computed the comparisons of CED curves on various training strategies with respect to our STRRN. The partition method for each facial components is consistent with that was defined in our spatial module. For the baseline method, we selected the detector HGN~\cite{DBLP:conf/eccv/NewellYD16} that was fine-tuned by our STRRN. Table~\ref{tab:ts} demonstrates the CED comparisons and
two-fold conclusions can be made from Table~\: 1) $\lambda$=0.4  achieves higher performance than  training by $\mathcal{L}_{\text{tck}}$ or $\mathcal{L}_{\text{det}}$ independently, and 2) $\mathcal{L}_{\text{tck}}$ is relatively important than $\mathcal{L}_{\text{det}}$ for robust spatial-temporal relation reasoning. Moreover, Table~\ref{tab:ts} also tabulates the comparisons of our STRRN versus independent spatial and temporal modules, which indicates the contributions of the complementary information of both modules.

\begin{table}
	\caption{Ablation experiments of our STRRN, where the 300VW dataset was employed for evaluation.} \small
	\vspace{5pt}\centering
	\begin{tabular}{ | l  | c | c | c |} 
		\hline
		\textbf{Method} & \textbf{Cate-1} &\textbf{ Cate-2 }& \textbf{ Cate-3}   \\
		\hline
		\textbf{STRRN}~(\textit{ w/o} $\mathcal{L}_{\text{tck}}$) &{7.92} & {{5.78}} & {8.78}  \\
		\textbf{STRRN}~(\textit{ w/o} $\mathcal{L}_{\text{det}}$, $\lambda$={0})  &{7.92} & {{7.08}} & {8.26} \\
		\textbf{STRRN}~({ $\mathcal{L}_{\text{tck}}$ and  $\mathcal{L}_{\text{det}}$,	$\lambda=$1.6 })    &{7.02} &{{6.91}}  & {8.21}  \\
		\hline
		\textbf{STRRN}~(\textit{w/o} Temporal Module)  & 8.76 &  7.98 & 8.93\\
		\textbf{STRRN}~(\textit{w/o} Spatial Module) & 8.45 &  7.84 & 8.76\\
		\hline
		\textbf{STRRN}~(S \& T Module, $\lambda=0.4$)  &\textbf{5.03} & {\textbf{4.74}} & \textbf{6.63}\\
		\hline
		
	\end{tabular}\label{tab:ts}
\end{table}

\textbf{Computational Efficiency:}  Our STRRN requires the computation complexity of $\mathcal{O}(N_R^2)$ for the geometry construction, where $N_R$ denotes the landmark scalability within each facial component. According to the 300-W annotation, $N_R$ maximally reaches approximate $20 \times (20+1)$ for the mouth region. 
For the efficient training, our architecture was implemented under the parallel-computing deep learning Tensorflow.  
The whole training procedure processes at about 60\textit{ms} each frame with a GPU of single NVIDIA GTX 1080 Ti graphic computation card~(11G memory).  Excluding the time of the face detection part, our model runs at 30 frames per second on one CPU with the Intel(R) Core(TM) i5-6500 CPU@3.20GHz and requires around 2G memory usage for runtime data loading.

\section{Conclusion}
In this work, we have investigated into the omni-supervised face alignment particular for large scale unlabeled video data. We have proposed a spatial-temporal relational reasoning networks method to reason about the spatial-temporal relation from the unlabeled videos. Extensive experimental results have showed that our approach have surpassed the performance of most state-of-the-art methods. How to apply reinforcement learning~\cite{DBLP:conf/eccv/BuchlerBO18,DBLP:journals/tpami/Liu19,DBLP:conf/eccv/GuoLZ18} to address the  3D face alignment~\cite{Jourabloo_2017_ICCV} with omni-supervision is desirable in our future works. 

\section*{Acknowledgment}
This work was supported in part by the National Science
Foundation of China under Grant 61806104, in part by the Natural Science Foundation of Ningxia under Grant 2018AAC03035, in part by
the Scientific Research Projects of Colleges and Universities
of Ningxia under Grant NGY2018050, and in part by the
Youth Science and Technology Talents Enrollment Projects
of Ningxia under Grant TJGC2018028

{\small
	\bibliographystyle{aaai}
	\bibliography{DCDRL,PAMI2018}
}

\end{document}